\documentclass[11pt]{article}

\usepackage[preprint]{acl}

\usepackage{times}
\usepackage{latexsym}
\usepackage[T1]{fontenc}
\usepackage[utf8]{inputenc}
\usepackage{microtype}
\usepackage{inconsolata}

\usepackage{booktabs}   
\usepackage{multirow}   

\usepackage{listings}

\usepackage{graphicx}

\usepackage{amsmath}

\usepackage{amssymb}
\usepackage{amsfonts}
\usepackage{amsthm}
\usepackage{mathtools}

\usepackage{booktabs}  

\usepackage{algorithm}
\usepackage{algorithmic}  

\usepackage{tcolorbox}
\tcbuselibrary{skins}
\usepackage{lipsum}
\usepackage{marvosym}

\theoremstyle{plain}
\newtheorem{theorem}{Theorem}[section]

\theoremstyle{definition}
\newtheorem{definition}[theorem]{Definition}

\theoremstyle{remark}

\newcommand{\Pre}{\mathrm{Pre}}
\newcommand{\Eff}{\mathrm{Eff}}

\newcommand{\Query}{\textsc{Query}}

\DeclareMathOperator*{\argmax}{arg\,max}

\title{Active Epistemic Control for Query-Efficient Verified Planning}

\author{Shuhui Qu \\
  Stanford University \\
  \texttt{shuhuiq@stanford.edu} \\
  }

\begin{document}
\maketitle

\begin{abstract}
Planning in interactive environments is challenging under partial observability: task-critical preconditions (e.g., object locations or container states) may be unknown at decision time, yet grounding them through interaction is costly. Learned world models can cheaply predict missing facts, but prediction errors can silently induce infeasible commitments. We present \textbf{Active Epistemic Control (AEC)}, an epistemic--categorical planning layer that integrates model-based belief management with categorical feasibility checks. AEC maintains a strict separation between a \emph{grounded fact store} used for commitment and a \emph{belief store} used only for pruning candidate plans. At each step, it either queries the environment to ground an unresolved predicate when uncertainty is high or predictions are ambiguous, or simulates the predicate to filter hypotheses when confidence is sufficient. Final commitment is gated by grounded precondition coverage and an SQ-BCP pullback-style compatibility check, so simulated beliefs affect efficiency but cannot directly certify feasibility. Experiments on ALFWorld and ScienceWorld show that AEC achieves competitive success with fewer replanning rounds than strong LLM-agent baselines.
\end{abstract}

\section{Introduction}
\label{sec:intro}

Reliable planning under partial observability remains a central challenge for interactive agents, especially when correctness depends on latent preconditions that must be verified through costly interaction \cite{ghallab2004automated,wang2022scienceworld,shridhar2020alfworld}. In embodied benchmarks such as ALFWorld \cite{shridhar2020alfworld}, even seemingly simple goals (e.g., \emph{cool an apple and place it in the microwave}) hinge on predicates such as \texttt{apple\_location}, \texttt{apple\_temperature}, and \texttt{microwave\_open} that are often not grounded from the initial observation. Agents thus face a recurring trade-off: \emph{query} the environment to ground missing facts (reliable but expensive), or \emph{simulate} them with a learned world model (cheap but error-prone). When this trade-off is mishandled, agents either (i) succeed less often due to ungrounded assumptions, or (ii) incur more replanning by repeatedly repairing failed plans.

Large Language Models (LLMs) \cite{achiam2023gpt,grattafiori2024llama,touvron2023llama} offer strong generative capabilities for planning and tool use \cite{wei2022chain,yao2022react}, but frequently violate environment-specific constraints when critical preconditions are unobserved \cite{kambhampati2024llms,valmeekam2023planning}. Classical symbolic planning provides formal correctness \cite{ghallab2004automated,helmert2006fast}, but typically assumes a fully specified state and requires substantial domain engineering to remain usable in open-ended interactive settings. Recent agentic methods improve reasoning and recovery through interaction and self-correction \cite{yao2022react,shinn2023reflexion}, yet they do not make explicit \emph{when} verification is necessary, nor do they prevent model beliefs from implicitly becoming ``facts'' during commitment.

World-model-driven approaches offer an attractive middle ground, but differ sharply in how they treat verification. WKM \cite{qiao2024agent} relies on parametric prediction without online grounding, reducing interaction but risking silent failure when preconditions are hallucinated. WALL-E \cite{zhou2024wall} improves via failure-driven refinement and neurosymbolic rules, but still depends on a planning-time choice of what to ground and what to assume, often manifesting as additional replanning when assumptions are wrong. More broadly, these methods tie correctness tightly to model quality: stronger predictors help, but do not address the structural question of how to use predictions \emph{without} letting them certify feasibility.

We address this gap by introducing \textbf{Active Epistemic Control (AEC)}, a planning-time framework that unifies learned prediction with grounded verification. AEC maintains two explicitly separated stores: a grounded fact store $w$ that contains only externally supported predicate assignments, and a belief store $\hat{w}$ that records model predictions annotated with uncertainty. At each step, AEC either (i) queries the environment to ground an unresolved predicate when uncertainty is high or predictions are ambiguous, or (ii) simulates the predicate to prune candidate hypotheses when confidence is sufficient. Crucially, AEC commits plans only through a verifier that depends \emph{only} on grounded evidence, integrating sound grounded precondition coverage with a categorical compatibility check inherited from SQ-BCP \cite{qu2025sqbcp}. This yields a clear separation: simulation can affect efficiency (which hypotheses survive), but cannot directly certify feasibility.

We evaluate AEC on ALFWorld and report competitive success with reduced replanning and token cost under the standard metrics used in prior interactive-agent evaluations \cite{zhou2024wall}. We additionally include ScienceWorld \cite{wang2022scienceworld} as a second benchmark to assess generality beyond household tasks. Overall, AEC provides a principled control layer for deciding \emph{when} to verify under partial observability, while preserving grounded-only commitment in the presence of imperfect world models.

\paragraph{Contributions.}

\begin{enumerate}
\item \textbf{Epistemic–categorical integration.} We integrate categorical feasibility checks (pullbacks from SQ-BCP) into an epistemic controller that decides which missing preconditions to ground under uncertainty.
\item \textbf{Separation principle.} We formalize and operationalize a separation between (i) model-based simulation for hypothesis pruning and (ii) grounded evidence for commitment, ensuring that compositional feasibility is certified only from grounded facts.
\item \textbf{Empirical validation on embodied benchmarks.} Using WALL-E metrics, we show competitive success with reduced replanning, and we extend evaluation to ScienceWorld to test generality beyond ALFWorld.
\end{enumerate}

\section{Related Work}
\label{sec:related_work}

\paragraph{LLM-based planning and tool-using agents.}
Large language models (LLMs) are increasingly used as high-level planners and interactive controllers that map natural-language goals to action sequences. \citep{achiam2023gpt,touvron2023llama,grattafiori2024llama}
Prompting strategies such as Chain-of-Thought expose intermediate reasoning that can improve multi-step decision making. \citep{wei2022chain}
ReAct couples reasoning traces with action/tool calls, enabling agents to iteratively ground information through interaction. \citep{yao2022react}
Reflexion and related trial-and-revision paradigms leverage feedback from failures to refine subsequent attempts. \citep{shinn2023reflexion}
Despite strong empirical performance, LLM agents remain brittle when correctness depends on unobserved constraints or environment-specific dynamics that are not captured in the prompt. \citep{valmeekam2023planning,kambhampati2024llms,wang2024planning}
More broadly, evaluations of LLM planning highlight systematic feasibility and constraint-satisfaction failures under realistic settings. \citep{valmeekam2023planning,stein2023autoplanbench,wu2025haste}

\paragraph{World-model-driven planning and simulation.}
A parallel line of work reduces online interaction by predicting unobserved facts using learned world models. \citep{ha2018world,schrittwieser2020mastering}
In LLM-agent settings, world-model-centric approaches can lower interaction cost, but may fail silently when predicted preconditions are treated as truth. \citep{qiao2024agent, wang2024planning}
WKM exemplifies parametric prediction to support planning without explicit online verification of every missing predicate. \citep{qiao2024agent}
Programmatic world modeling adds an executable substrate for simulation: WorldCoder induces code-level environment models and plans against the induced simulator. \citep{tang2024worldcoder}
However, correctness in programmatic simulation remains sensitive to mismatches between induced dynamics and the runtime environment, especially under distribution shift. \citep{tang2024worldcoder,wang2024planning}
AEC is complementary: it treats world-model outputs as \emph{beliefs for pruning} and gates commitment on grounded-only verification, so model errors primarily affect efficiency rather than feasibility. \citep{qiao2024agent,tang2024worldcoder}

\paragraph{Rule learning and failure-driven refinement.}
Neurosymbolic agents increasingly incorporate induced rules or constraints to override unreliable predictions and enforce discovered environment regularities. \citep{zhou2024wall}
WALL-E is a representative failure-driven refinement approach that learns override rules from counterexamples collected during interaction and iterates to improve performance. \citep{zhou2024wall}
Such refinement loops are effective, but they still face the decision problem of which beliefs must be grounded \emph{before} committing to a plan in a new episode. \citep{zhou2024wall,kaelbling1998planning}
AEC can reuse the same kind of trial feedback, but our emphasis is the \emph{control layer} that arbitrates between querying (to ground facts) and simulating (to cheaply filter hypotheses) while maintaining a grounded-only commitment story. \citep{zhou2024wall,kaelbling1998planning}

\paragraph{Active information gathering under uncertainty.}
AEC is closely related to planning under partial observability, where an agent must trade off information acquisition cost against the risk of acting on incorrect beliefs. \citep{kaelbling1998planning,albore2009translation}
Classical POMDP and contingent-planning formulations provide principled treatments of epistemic actions, but exact solutions are often intractable and require accurate models. \citep{kaelbling1998planning,albore2009translation}
In contrast, many LLM-agent methods rely on implicit uncertainty (e.g., self-consistency) or ad hoc heuristics for verification decisions. \citep{wei2022chain,yao2022react,kambhampati2024llms}
AEC makes uncertainty operational as an explicit control signal (query vs.\ simulate) while preventing belief leakage by conditioning predictions only on grounded facts. \citep{kaelbling1998planning}

\paragraph{Structured verification and compositional planning.}
Symbolic planning provides formal guarantees when the domain model and state are correct, but typically requires substantial domain engineering and struggles with missing or noisy state. \citep{ghallab2004automated,helmert2006fast}
Recent neurosymbolic approaches introduce explicit verifiers or constraint checkers to filter candidate plans produced by neural components. \citep{valmeekam2023planning,wang2024planning}
AEC’s verifier combines grounded precondition coverage (via sound-but-incomplete entailment) with a categorical compatibility check inherited from SQ-BCP, so plan commitment depends only on grounded evidence. \citep{qu2025sqbcp}
This positions AEC as a lightweight safety gate that can sit atop LLM planners and learned world models without requiring predicted beliefs to be treated as truth. \citep{qu2025sqbcp,kaelbling1998planning}

\section{Active Epistemic Control Framework}
\label{sec:aec}

\subsection{Background: Category-Theoretic Planning}
We build on the categorical planning formalism introduced in SQ-BCP~\citep{qu2025sqbcp}. A planning problem is represented as a category whose objects are world states and whose morphisms are actions (or action compositions). This perspective isolates (i) \emph{structural feasibility} from (ii) the \emph{epistemic} challenge induced by partial observability.

In our benchmarks, a world state is represented by a set of typed predicate facts. A candidate plan (hypothesis) $h$ is a composed morphism annotated with (a) a precondition set $\Pre(h)$ and (b) an expected binary assignment $h.\mathrm{expected}(p)\in\{0,1\}$ for each $p\in\Pre(h)$ (as produced by the planner/annotator). AEC does not require a particular hypothesis generator: any procedure that outputs a finite set of candidates $H$ with $\Pre(h)$ and $h.\mathrm{expected}(\cdot)$ is compatible.

\subsection{Epistemic State: Grounded Facts vs.\ Beliefs}
AEC maintains an explicit separation between \emph{grounded} (externally supported) facts and \emph{beliefs} (model predictions) used only for hypothesis pruning.

\begin{definition}[Epistemic State]
\label{def:epistemic_state}
An epistemic state is a triple $s=(w,\hat{w},H)$, where $w$ is a set of grounded predicate assignments, $\hat{w}$ is a belief store of tuples $(p,\hat{v}_p,\sigma_p)$, and $H$ is a set of candidate hypotheses (plans).
\end{definition}

\paragraph{Grounded store.}
$w$ contains only facts obtained from (i) the initial observation $w_0$ (grounded but partial), and (ii) subsequent environment interactions that provide verifiable evidence (queries and/or execution feedback). Unobserved predicates are not assumed false; they remain unknown until grounded.

\paragraph{Belief store.}
$\hat{w}$ contains tentative predictions only and is never treated as truth. Each entry $(p,\hat{v}_p,\sigma_p)$ records a discretized belief $\hat{v}_p\in\{0,1\}$ together with an epistemic uncertainty estimate $\sigma_p\in[0,1]$.

\paragraph{Unresolved preconditions.}
Let $\mathrm{dom}(w)=\{p\mid (p,v)\in w\}$ and $\mathrm{dom}(\hat{w})=\{p\mid (p,\hat{v},\sigma)\in\hat{w}\}$. For any hypothesis $h$, we write the unresolved set as
\begin{equation}
U(w,\hat{w},h)=\{p\in \Pre(h)\mid p\notin \mathrm{dom}(w)\cup \mathrm{dom}(\hat{w})\}.
\end{equation}
AEC repeatedly selects $p\in U(\cdot)$ to resolve until either all remaining candidates are verifiable or the query budget is exhausted.

\paragraph{No belief leakage (key invariance).}
All predictions are conditioned \emph{only} on grounded facts $w$, never on beliefs $\hat{w}$. This prevents self-confirming feedback loops in which earlier model guesses become inputs that reinforce later guesses.

\subsection{Queries and Grounded Updates}
AEC resolves unknown predicates via (i) external queries that ground facts, or (ii) world-model simulation that creates beliefs used only for hypothesis filtering.

\paragraph{Query macro-action and side effects.}
A query $\Query(p)$ is a macro-action that verifies predicate $p$ through environment interaction (possibly multiple primitive steps), parses the resulting observation to obtain $v\in\{0,1\}$, and returns any additional grounded facts revealed after the query. Importantly, queries need not be reversible (e.g., opening a container changes the environment). Therefore, after querying $p$ we update the grounded store as
\begin{equation}
\label{eq:query}
\begin{split}
    (v,\Delta w)&\leftarrow \Query(p)\\
    w &\leftarrow w\cup\{(p,v)\}\cup \Delta w.
\end{split}
\end{equation}
This ensures $w$ remains a sound representation of all grounded facts available to the agent.

\subsection{Predictive Model Interface and Ambiguity Margin}
AEC requires a predictor that estimates both a belief score and epistemic uncertainty:
\begin{equation}
M_\theta:(w,p)\mapsto(\mu_p,\sigma_p),
\end{equation}
where $\mu_p\in[0,1]$ estimates whether predicate $p$ holds given grounded facts $w$, and $\sigma_p\in[0,1]$ is an epistemic uncertainty estimate. 

The AEC controller uses $M_\theta$ only through $(\mu_p,\sigma_p)$ and is otherwise agnostic to the specific uncertainty estimator (e.g., ensembles, Bayesian estimators). In practice, calibrated ensemble uncertainty provides a simple choice under partial observability. In our experiments, $M_\theta$ follows the same LLM-based predictive interface used in WALL-E\citep{zhou2024wall} with ensemble output.

\paragraph{Discretization with margin.}
To avoid brittle switching near the decision boundary, AEC discretizes a prediction only when it is sufficiently far from $0.5$:
\begin{equation}
\hat{v}_p=\mathbb{1}[\mu_p\ge 0.5]
\quad \text{only if}\quad |\mu_p-0.5|\ge \epsilon.
\end{equation}
If $|\mu_p-0.5|<\epsilon$, the prediction is treated as ambiguous and is resolved by querying.

\subsection{Verification: Grounded Coverage + Categorical Consistency}
AEC commits to a plan only if it passes a verifier that depends only on grounded facts (and sound entailment from them).

\begin{definition}[Sound Verifier]
\label{def:verifier}
A verifier $\mathcal{V}(h,w,w^\star)\in\{\textsc{True},\textsc{False}\}$ is sound if

\begin{equation}
\begin{split}
    &\mathcal{V}(h,w,w^\star)=\textsc{True}\ \Rightarrow\ \\
    &\begin{cases}
\forall p\in\Pre(h): \big(p\in\mathrm{dom}(w)\big)\ \lor\ \big(w\models p\big),\\
\mathrm{Apply}(h,w)\ \text{satisfies goal constraints in}\ w^\star,
\end{cases}
\end{split}
\end{equation}
where $w\models p$ denotes entailment via a fixed set of sound (but possibly incomplete) inference rules, and $\mathrm{Apply}(h,w)$ denotes symbolic execution of $h$ from $w$ (Definition~\ref{def:apply}).
\end{definition}

\begin{definition}[Plan Application]
\label{def:apply}
Let $h$ be a composed morphism (plan) with action sequence $(f_1,\dots,f_T)$. We define
\[
\mathrm{Apply}(h,w) \;=\; \Eff(f_T)\circ\cdots\circ\Eff(f_1)(w),
\]
i.e., the symbolic state obtained by applying the effects of each action in sequence.
\end{definition}

\paragraph{Instantiation.}
We implement $\mathcal{V}$ as the conjunction of:
(i) a \emph{precondition coverage check} (grounded-or-entailed), and
(ii) a \emph{categorical pullback check} (SQ-BCP) for compositional compatibility:
\begin{equation}
\begin{split}
    \mathcal{V}(h,w,w^\star)=&\textsc{CheckPre}(h,w)\ \land\ \\ &\textsc{PullbackVerify}(h,w,w^\star).
\end{split}
\end{equation}
The entailment rules are \emph{sound but incomplete}: they never assert false implications, but may fail to derive true ones, leading to additional queries or conservative rejections (efficiency loss only, not correctness).

\subsection{Uncertainty-Guided Epistemic Control}
Algorithm~\ref{alg:aec} presents the AEC policy. At each iteration, it selects a predicate $p^\star\in U$, predicts $(\mu,\sigma)=M_\theta(w,p^\star)$ using grounded facts only, then either (i) queries (grounding $p^\star$ and updating $w$), or (ii) simulates (recording $(p^\star,\hat{v},\sigma)$ in $\hat{w}$ and pruning $H$). Finally, it commits to the best surviving hypothesis only if it passes verification on grounded facts.

\begin{algorithm}[t]
\caption{Uncertainty-Guided Epistemic Control (AEC)}
\label{alg:aec}
\begin{algorithmic}[1]
\REQUIRE Initial grounded state $w_0$, goal constraints $w^\star$, world model $M_\theta$, threshold $\tau$, margin $\epsilon$
\ENSURE Verified plan $\pi$ or \textsc{Fail}
\STATE $H \leftarrow \textsc{GenerateHypotheses}(w_0,w^\star)$
\STATE $w \leftarrow w_0$ \COMMENT{grounded facts (partial)}
\STATE $\hat{w}\leftarrow\emptyset$ \COMMENT{belief store}
\STATE $U \leftarrow \bigcup_{h\in H} U(w,\hat{w},h)$
\STATE $q_{\mathrm{budget}}\leftarrow \textsc{MaxQueries}$
\WHILE{$|U|>0$ \AND $q_{\mathrm{budget}}>0$ \AND $|H|>0$}
    \STATE $p^\star\leftarrow \textsc{SelectPrecondition}(U,H)$
    \STATE $(\mu,\sigma)\leftarrow M_\theta(w,p^\star)$ 
    \IF{$|\mu-0.5|<\epsilon$} 
        \STATE $(v,\Delta w)\leftarrow \Query(p^\star)$
        \STATE $w \leftarrow w \cup \{(p^\star,v)\}\cup \Delta w$
        \STATE $H \leftarrow \{h\in H\mid h.\mathrm{expected}(p^\star)=v\}$
        \STATE $q_{\mathrm{budget}}\leftarrow q_{\mathrm{budget}}-1$
    \ELSIF{$\sigma>\tau$} 
        \STATE $(v,\Delta w)\leftarrow \Query(p^\star)$
        \STATE $w \leftarrow w \cup \{(p^\star,v)\}\cup \Delta w$
        \STATE $H \leftarrow \{h\in H\mid h.\mathrm{expected}(p^\star)=v\}$
        \STATE $q_{\mathrm{budget}}\leftarrow q_{\mathrm{budget}}-1$
    \ELSE
        \STATE $\hat{v}\leftarrow \mathbb{1}[\mu\ge 0.5]$
        \STATE $\hat{w}\leftarrow \hat{w}\cup \{(p^\star,\hat{v},\sigma)\}$
        \STATE $H \leftarrow \{h\in H\mid h.\mathrm{expected}(p^\star)=\hat{v}\}$
    \ENDIF
    \STATE $U \leftarrow \bigcup_{h\in H} U(w,\hat{w},h)$
\ENDWHILE
\STATE $h^\star \leftarrow \argmax_{h\in H}\textsc{Score}(h)$
\IF{$\mathcal{V}(h^\star,w,w^\star)=\textsc{True}$} 
    \STATE \textbf{return} plan induced by $h^\star$
\ELSE
    \STATE \textbf{return} \textsc{Fail}
\ENDIF
\end{algorithmic}
\end{algorithm}

\subsection{Correctness: Simulation Affects Efficiency, Not Feasibility}
AEC maintains a grounded store $w$ and a belief store $\hat{w}$. Querying may add facts to $w$, while simulation
adds entries only to $\hat{w}$. Crucially, plan commitment is gated by a verifier $\mathcal{V}$ that depends only on
grounded facts $w$ (and not on $\hat{w}$). Therefore, simulation can affect which candidate plan survives, but cannot
directly certify feasibility.

\begin{theorem}[Verified Epistemic Control]
\label{thm:verified_final}
Consider one run of Algorithm~\ref{alg:aec}. Let $w_0$ be the initial grounded facts and let
$w_{\mathrm{final}}$ be the grounded facts at the moment a plan $\pi$ is committed. Define
\begin{equation}
Q \;:=\; \mathrm{dom}(w_{\mathrm{final}})\setminus \mathrm{dom}(w_0),
\end{equation}
i.e., the set of predicates whose truth values were grounded \emph{after} initialization. Assume:
\begin{enumerate}
    \item \textbf{Grounded-only verification:} $\pi$ is committed only if
    $\mathcal{V}(\pi,w_{\mathrm{final}},w^\star)=\textsc{True}$, and $\mathcal{V}$ depends only on $w_{\mathrm{final}}$
    (not on $\hat{w}$).
    \item \textbf{Verifier soundness:} if all facts in $w_{\mathrm{final}}$ are correct and
    $\mathcal{V}(\pi,w_{\mathrm{final}},w^\star)=\textsc{True}$, then $\pi$ is feasible.
    \item \textbf{Query error model:} for each $p\in Q$, the probability that the grounded value of $p$ in
    $w_{\mathrm{final}}$ is incorrect is at most $\varepsilon_{\mathrm{oracle}}(p)$.
\end{enumerate}
Then the committed plan $\pi$ satisfies
\begin{equation}
\Pr(\pi\ \text{is feasible}) \;\ge\; 1-\sum_{p\in Q}\varepsilon_{\mathrm{oracle}}(p).
\end{equation}
In particular, errors made by simulation-only predicates (those appearing only in $\hat{w}$) do not enter the bound.
\end{theorem}

\paragraph{Proof sketch.}
Simulation never writes into $w$, and the verifier reads only $w$. Thus, if all facts in $w_{\mathrm{final}}$ are correct,
verifier soundness implies $\pi$ is feasible. The only facts in $w_{\mathrm{final}}$ that can be incorrect are those grounded
after initialization, namely predicates in $Q$. By the query error model, each $p\in Q$ is incorrect with probability at most
$\varepsilon_{\mathrm{oracle}}(p)$. Therefore, by a union bound,
\[
\Pr(\exists\, p\in Q\ \text{s.t. $p$ is incorrect}) \;\le\; \sum_{p\in Q}\varepsilon_{\mathrm{oracle}}(p),
\]
so with probability at least $1-\sum_{p\in Q}\varepsilon_{\mathrm{oracle}}(p)$, all facts in $w_{\mathrm{final}}$ are correct,
and hence $\pi$ is feasible.

\paragraph{Practical note: failure-driven refinement}
The guarantee above is \emph{per run}: it relies only on grounded-only commitment and the query error model, and is independent of how the world model is trained. In our implementation, we optionally reuse success/failure trajectories \emph{across iterations} to improve efficiency: counterexamples from failed episodes are aggregated to (i) update the backbone predictor or its calibration for precondition uncertainty, and (ii) patch verifier mismatches when a grounded query contradicts an entailed predicate. These updates change future query allocation and replanning frequency, but do not alter the fact that commitment in any given run is gated by $\mathcal{V}$ on grounded evidence.

\section{Experiments}
\label{sec:experiments}

We evaluate Active Epistemic Control (AEC) on embodied planning tasks under partial observability. Our evaluation focus on (i) task completion, (ii) replanning behavior under failure, and (iii) LLM token usage. Our primary goal is to measure how grounded verification and uncertainty-guided querying affect success and replanning efficiency, without relying on iterative multi-episode training at test time.

\subsection{Benchmarks}
\label{subsec:benchmarks}

\paragraph{ALFWorld.}
ALFWorld~\citep{shridhar2020alfworld} is a household task benchmark in which an agent must complete language-conditioned goals (e.g., \emph{cool an apple then place it in the microwave}) via discrete actions under partial observability. Episodes have a fixed interaction budget (50 steps in our setup). Many task-critical predicates (object locations, container states, object attributes such as clean/hot/cool) are not fully observable initially and must be resolved through interaction. Following prior work, we evaluate on the standard set of 134 predefined test tasks.

\paragraph{ScienceWorld (planned).}
ScienceWorld~\citep{wang2022scienceworld} is an interactive science reasoning environment from AllenAI featuring long-horizon tasks (e.g., conducting experiments, manipulating objects, controlling environment conditions) under partial observability. ScienceWorld includes richer state transitions and more complex preconditions (e.g., temperature, phase changes, containment, device operation), making it a suitable stress test for AEC's separation of grounded facts and beliefs as well as its verifier-based commitment rule.
We will report results on a held-out test set following a standard train/test split used in prior ScienceWorld agent evaluations; details will be finalized with the released evaluation protocol and provided in the camera-ready version.

\subsection{Evaluation Metrics}
\label{subsec:metrics}

We report the following metrics

\paragraph{Success rate (\%, higher is better).}
The percentage of tasks completed successfully within the episode budget.

\paragraph{Replanning rounds (lower is better).}
The number of times the agent re-enters a planning phase for the same task due to execution failure or contradiction (e.g., an action fails, a queried predicate contradicts a plan precondition, or the verifier rejects the current best hypothesis). If the task is not completed, replanning rounds are set to a maximal cap.

\subsection{Baselines}
\label{subsec:baselines}

We compare against representative LLM-based planning and world-modeling approaches. All methods are evaluated with the same environment API and observation interface.

\begin{itemize}
    \item \textbf{Direct}: A direct LLM planner that proposes a plan from the task description and current observation, without explicit grounded verification.
    \item \textbf{ReAct}~\citep{yao2022react}: Interleaves reasoning and acting; replanning occurs naturally through interaction loops.
    \item \textbf{WKM}~\citep{qiao2024agent}: A world-knowledge model that predicts missing state predicates without explicit grounded verification.
    \item \textbf{WALL-E}~\citep{zhou2024wall}: Failure-driven refinement via learned neurosymbolic rules and MPC-style monitoring. We report results using the released codebase under the same evaluation caps as AEC.
    \item \textbf{Query-Only}: An upper-bound variant that resolves all missing predicates through queries before committing to a plan (interaction- and token-heavy, but intended to reduce state uncertainty).
\end{itemize}

\subsection{AEC Configuration}
\label{subsec:aec_config}

\paragraph{Hypothesis generation.}
AEC requires a finite set of candidate plans $H$. We generate a small set of diverse hypotheses from the initial grounded facts $w_0$ and goal constraints $w^\star$ using a single LLM planner. Each hypothesis includes an extracted precondition set $\mathrm{Pre}(h)$ and expected binary assignments $h.\mathrm{expected}(p)$.

\paragraph{World model interface.}
AEC uses a predictive model $M_\theta$ that maps $(w,p)$ to a belief score $\mu_p$ and uncertainty $\sigma_p$. The controller conditions only on grounded facts $w$ (no belief leakage), and discretizes predictions using an ambiguity margin $\epsilon$ (Section~\ref{sec:aec}). We use wall-E@5 as the world model

\paragraph{Query policy.}
AEC queries predicates that are either ambiguous ($|\mu_p - 0.5| < \epsilon$) or high-uncertainty ($\sigma_p > \tau$). We use a fixed $\epsilon$ across all experiments and select $\tau$ on a validation set.

\paragraph{Verifier.}
AEC commits to a plan only if it passes a grounded verifier $\mathcal{V}$ consisting of (i) a precondition coverage check using grounded facts and sound entailment and (ii) a pullback-based compositional consistency check (Section~\ref{sec:aec}).

\section{Results}
\label{sec:results}

\subsection{Main Results on ALFWorld}
\label{subsec:main_results_alfworld}

\begin{table*}[t]
\centering
\small
\setlength{\tabcolsep}{6pt}
\begin{tabular}{clccccccc}
\toprule
\multicolumn{2}{c}{\textbf{Method}} &
\multicolumn{7}{c}{\textbf{Success Rate (\%) $\uparrow$}} \\
\cmidrule(lr){3-9}
& & \textbf{Avg.} & \textbf{Pick} & \textbf{Clean} & \textbf{Heat} & \textbf{Cool} & \textbf{Examine} & \textbf{Picktwo} \\
\midrule

\multirow{5}{*}{\rotatebox{90}{\textbf{VLMs}}}
& MiniGPT-4* (Zhu et al., 2023a)          & 16 & 4  & 0  & 19 & 17 & 67 & 6  \\
& BLIP-2* (Li et al., 2023)              & 4  & 0  & 6  & 4  & 11 & 6  & 0  \\
& LLaMA-Adapter* (Gao et al., 2023b)     & 13 & 17 & 10 & 27 & 22 & 0  & 0  \\
& InstructBLIP* (Dai et al., 2023)       & 22 & 50 & 26 & 23 & 6  & 17 & 0  \\
& EMMA* (Yang et al., 2024)              & \textbf{82} & \textbf{71} & \textbf{94} & \textbf{85} & \textbf{83} & \textbf{88} & \textbf{67} \\
\midrule

\multirow{8}{*}{\rotatebox{90}{\textbf{LLMs}}}
& BUTLER* (Micheli \& Fleuret, 2021)     & 26 & 31 & 41 & 60 & 27  & 12  & 29 \\
& GPT-BUTLER* (Micheli \& Fleuret, 2021) & 69 & 62 & 81 & 85 & 78  & 50  & 47 \\
& DEPS (Wang et al., 2023a)              & 76 & 93 & 50 & 80 & \textbf{100} & \textbf{100} & 0  \\
& AutoGen* (Wu et al., 2023)             & 77 & 92 & 74 & 78 & 86  & 83  & 41 \\
& ReAct (Yao et al., 2023)               & 74 & 79 & 54 & 96 & 85  & 83  & 51 \\
& AdaPlanner (Sun et al., 2024)          & 91 & \textbf{100} & \textbf{100} & 89 & \textbf{100} & 97 & 47 \\
& Reflexion (Shinn et al., 2024)         & 86 & 92 & 94 & 70 & 81  & 90  & 88 \\
& RAFA (Liu et al., 2023)                & \textbf{95} & \textbf{100} & 97 & 91 & 95 & \textbf{100} & 82 \\
& WALL-E                          & \textbf{95} & \textbf{100} & 97 & \textbf{100} & 86 & 85 & \textbf{100} \\
& WALL-E2.0                          & \textbf{98.3} & \textbf{100} & 100 & \textbf{96} & 100 & 100 & \textbf{94} \\
& \textbf{AEC}            & 98.7 & 100 & 100 & 97 & 95 & 100 & 100 \\

\midrule
\multicolumn{2}{l}{\textbf{Human Performance}* (Shridhar et al., 2020a)} & 91 & -- & -- & -- & -- & -- & -- \\
\bottomrule
\end{tabular}

\caption{\textbf{Comparison of WALL-E and baselines on 134 testing tasks from the ALFWorld benchmark.}
* reported in previous work. VLMs = vision-language models, LLMs = large language models.
Success rate (\%) is the percentage of tasks completed successfully. The best score for each task is highlighted in \textbf{bold}.}
\label{tab:alfworld_wall_e_comparison}
\end{table*}

Table~\ref{tab:alfworld_wall_e_comparison} reports success rates on the 134 ALFWorld test tasks. Overall, AEC performs comparably to the strongest prior agents. AEC achieves \textbf{98.7\%} average success, which is close to WALL-E2.0 (98.3\%) and above earlier LLM baselines such as WALL-E/RAFA (95\%). We view this as evidence that the proposed epistemic control and grounded-only commitment can reduce a small number of residual failures even in a saturated regime.

While the absolute gains are modest, AEC attains robust performance across categories without relying on additional environment-specific metrics, supporting the claim that separating grounded facts from beliefs and gating commitment with verification is a reliable design choice for embodied planning under partial observability.

\subsection{Results on ScienceWorld}
\label{subsec:main_results_scienceworld}

\begin{table}[t]
\centering
\small
\begin{tabular}{lccc}
\toprule
\textbf{Llama3-8B} & \textbf{Seen} & \textbf{Unseen}  & \\
\midrule
ReAct& 24.76 & 22.66  \\
Reflexion & 27.23  & 25.41   \\
NAT & 55.24 &  48.76  \\
ETO & 57.90 &  52.33  \\
KnowAgent & 58.67 & 49.18   \\
WKM& 60.12 &  54.75  \\
\midrule
\textbf{AEC} & 61.36 &  58.62  \\
\bottomrule
\end{tabular}
\caption{ScienceWorld results.}
\label{tab:main_results_scienceworld}
\end{table}

Table~\ref{tab:main_results_scienceworld} reports results on ScienceWorld. We follow the WKM and split into \emph{Seen} and \emph{Unseen} tasks. AEC achieves \textbf{61.36} on Seen and \textbf{58.62} on Unseen, outperforming prior agents in Table~\ref{tab:main_results_scienceworld}. The gains are larger on \emph{Unseen} (+3.87 over WKM), suggesting that AEC’s grounded-first epistemic control improves robustness when the environment distribution shifts. The result shows that AEC’s separation of grounded facts from model beliefs and verifier-gated commitment provides a general mechanism for mitigating compounding state-inference errors in partially observed, multi-step tasks.

\subsection{Ablation Study}
\label{subsec:ablation_results}

\subsubsection{Components}
To isolate the contributions of AEC components, we ablate two core mechanisms of AEC: (i) the verifier (``w/o verification'' removes $\mathcal{V}$ and commits based on model-filtered hypotheses only), and (ii) the epistemic gating policy (``w/o gating'' disables the uncertainty/ambiguity-based gating used to decide when to query vs.\ simulate, while keeping the rest of the pipeline unchanged).

\begin{figure}[t]
    \centering
    \includegraphics[width=\linewidth]{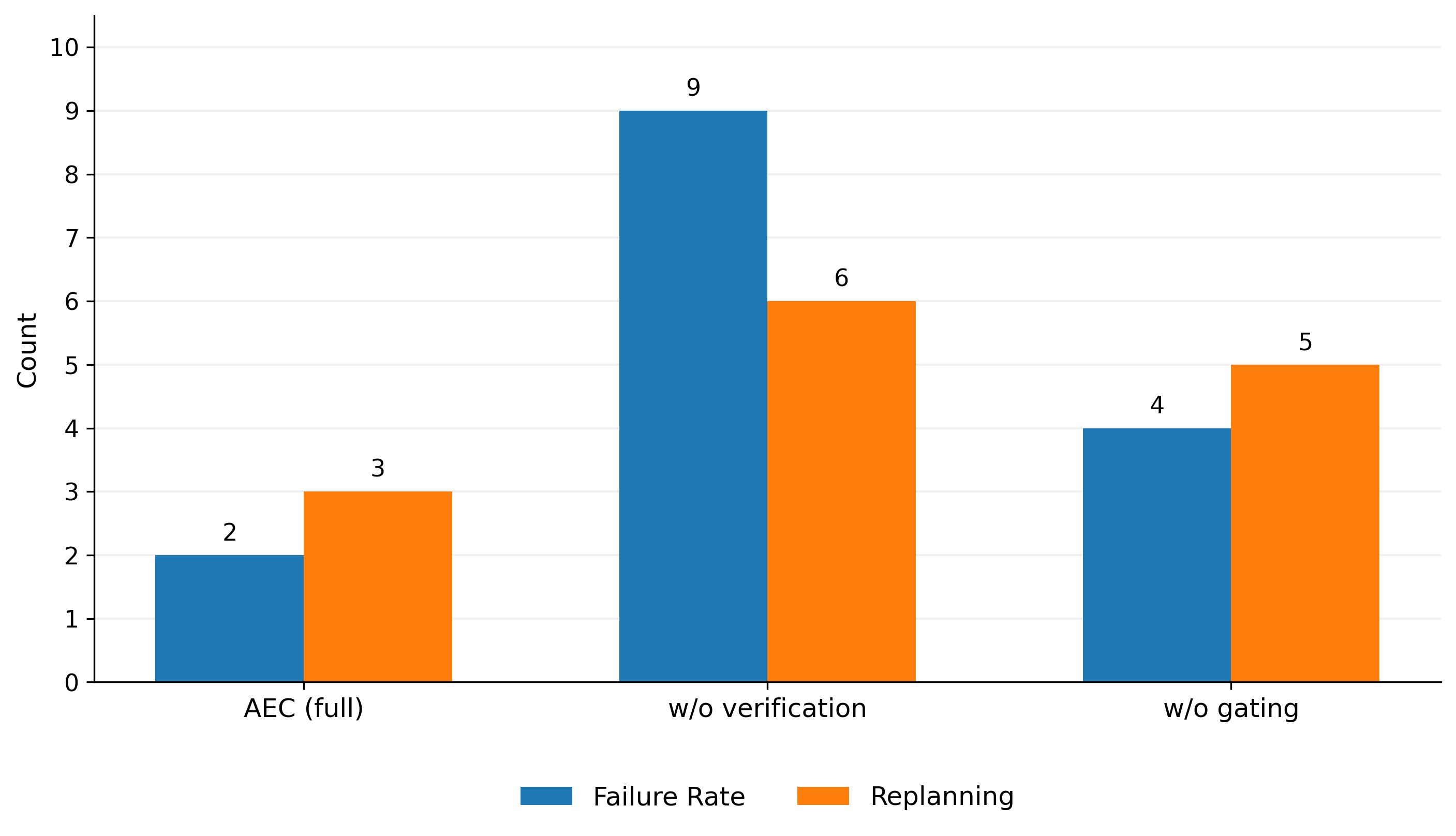}
    \caption{Ablation on ALFWorld. Blue: failure rate. Orange: replanning rounds.}
    \label{fig:ablation}
\end{figure}

Figure~\ref{fig:ablation} summarizes the ablation on ALFWorld. The full AEC achieves a low failure rate (2) and requires fewer replanning rounds (3). Removing verification substantially degrades robustness: failure rate increases to ~9 and replanning rises to ~6, indicating that model-based hypothesis filtering alone is insufficient to reliably prevent infeasible commitments. Disabling gating yields a milder but consistent regression (failure rate ~4; replanning ~5), suggesting that uncertainty/ambiguity gating primarily stabilizes execution by avoiding overconfident simulation and reducing downstream plan revision.

\begin{figure}[t]
    \centering
    \includegraphics[width=\linewidth]{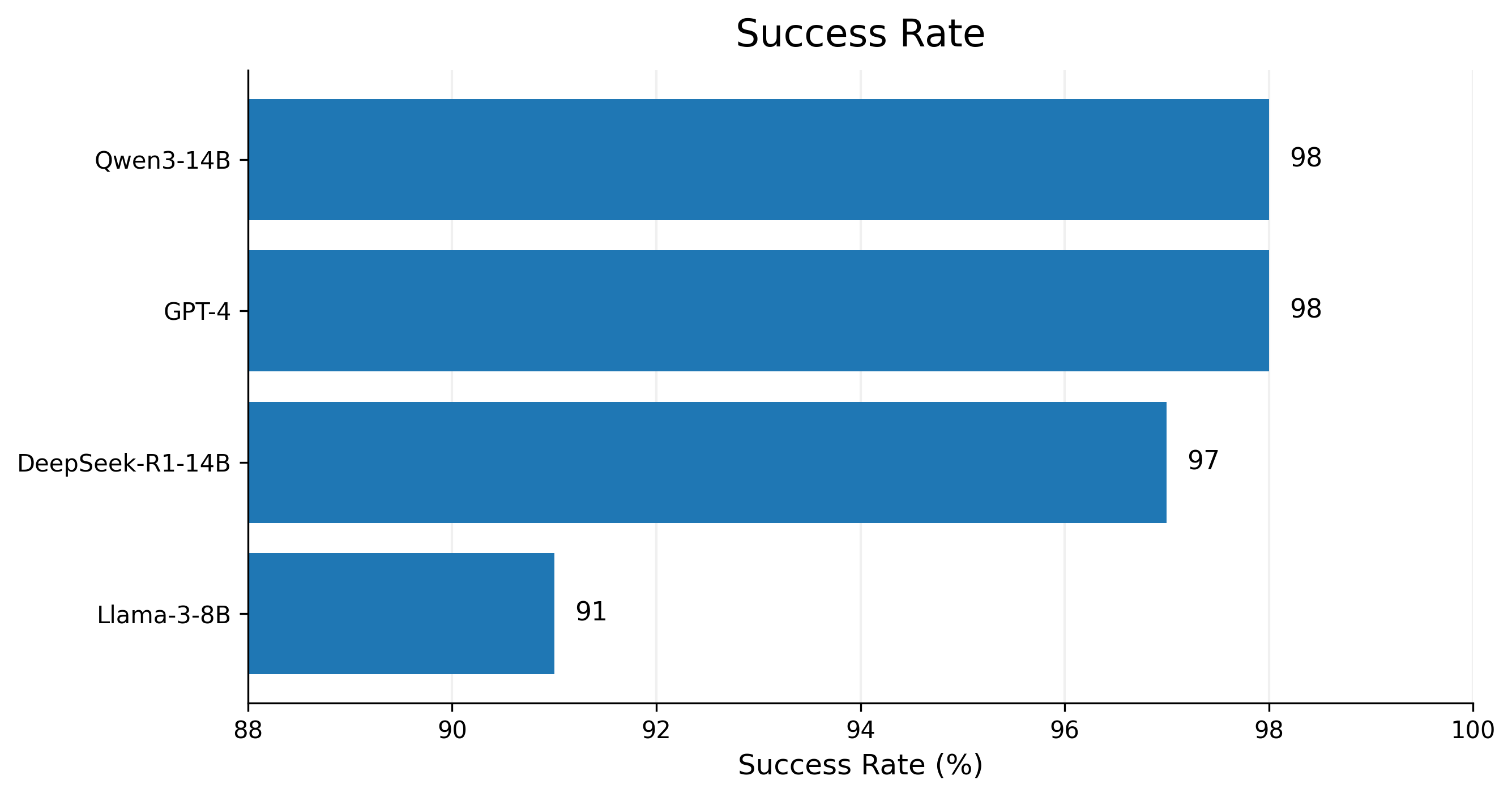}
    \caption{ALFWorld success rate of AEC under different LLM backbones.}
    \label{fig:backbone}
\end{figure}

\paragraph{Backbone sensitivity.}
We further evaluate AEC under different LLM backbones while keeping the overall pipeline (hypothesis generation, gating, and grounded verification) unchanged. Figure~\ref{fig:backbone} shows that AEC remains effective across backbones: GPT-4 and Qwen3-14B achieve the highest success (98\%), DeepSeek-R1-14B is slightly lower (97\%), and Llama-3-8B exhibits a larger drop (91\%). Overall, these results suggest that AEC’s gains are not tied to a single proprietary model; however, stronger backbones still translate into higher success, consistent with the role of the LLM planner/world model in generating and filtering candidate hypotheses.

\subsubsection{Number of iterations.}
\begin{figure}[t]
    \centering
    \includegraphics[width=\linewidth]{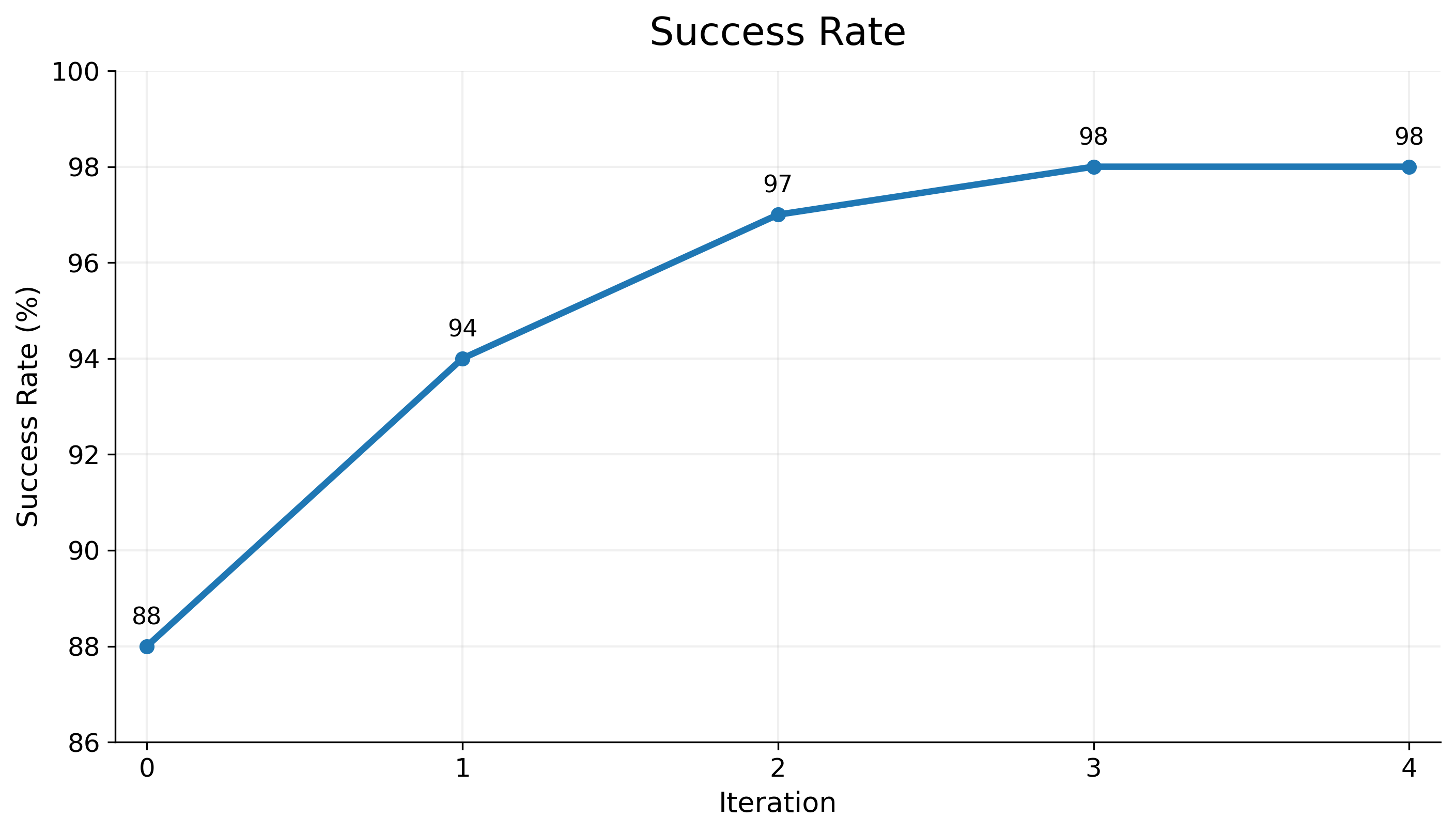}
    \caption{ALFWorld success rate of AEC across iterations.}
    \label{fig:iterations}
\end{figure}

Figure~\ref{fig:iterations} shows the effect of running AEC with failure-driven refinement across iterations. Success increases from $88\%$ at iteration~0 to $\approx 98\%$ by iterations~3--4, with most gains realized in the first two iterations. This indicates that the dominant failure modes are corrected early, and further iterations provide diminishing returns. We attribute the improvement to the refinement procedure described in Section~\ref{sec:aec}.

\section{Conclusion}
\label{sec:conclusion}

We presented \textbf{Active Epistemic Control (AEC)}, a planning framework for partially observable environments that explicitly separates \emph{grounded facts} from \emph{model beliefs}. AEC uses a predictive world model to cheaply prune candidate hypotheses, but commits to a plan only through a verifier that depends solely on grounded evidence (with sound-but-incomplete entailment and categorical consistency checks). This design yields a clear separation of concerns: simulation improves \emph{efficiency} by guiding which predicates to resolve, while feasibility is protected by grounded-only verification. Empirically, AEC achieves strong performance on ALFWorld and shows consistent gains over representative LLM-agent baselines, while remaining compatible with failure-driven refinement across iterations.

\paragraph{Limitations and future work.}
AEC’s guarantees are mediated by the grounding and verification interfaces: if the observation grounder or query oracle exhibits systematic errors (e.g., consistent visual misreads), AEC may incur conservative failures, extra replanning, or increased token cost even when a feasible plan exists. Moreover, our entailment rules are intentionally incomplete, which can inflate the set of unresolved predicates and shift work from logical inference to querying. Finally, AEC currently treats verification, replanning, and token usage as loosely coupled knobs rather than a single optimized objective.

These limitations suggest several directions. First, strengthen the verifier/grounder via counterexample-driven repair, including automatically detecting contradiction patterns and refining entailment rules or calibration thresholds from failure traces. Second, incorporate tighter decision objectives that jointly account for success, replanning rounds, and token usage, potentially learning instance-adaptive query thresholds rather than a fixed $\tau$. Third, evaluate broader generalization across environments with different observability and action semantics and characterize when improvements arise from better uncertainty estimation versus better hypothesis generation.

\newpage

\bibliography{custom}

\appendix

\section{Multimodal Grounding and Query Implementations}
\label{app:multimodal_grounding}

\subsection{Observation Grounder: Sound but Incomplete}
AEC assumes an \emph{observation grounder} that maps raw observations to an initial grounded fact set:
\begin{equation}
\mathcal{G}:(o^{\mathrm{img}},o^{\mathrm{text}})\mapsto w_0.
\end{equation}
We enforce the following design constraint:

\paragraph{Soundness requirement.}
$\mathcal{G}$ must be \emph{sound}: it should not assert a predicate $(p,1)$ unless it is supported by direct observation evidence.
When evidence is insufficient, the grounder abstains (omits $p$), leaving it unknown.
This makes grounding \emph{incomplete} by construction, which can increase the number of unresolved preconditions but does not compromise correctness because plan commitment is gated by $\mathcal{V}$ using grounded evidence.

\paragraph{Predicate inventory.}
For each benchmark, we define a typed predicate schema (objects, relations, and attributes) consistent with the environment API.
In ALFWorld, predicates cover container states (open/closed), object locations (in/on relations), and object attributes (clean/hot/cool).
In VirtualHome, predicates additionally cover reachability and interaction constraints exposed by the simulator.

\subsection{Query Macro-Actions}
A query $\Query(p)$ is implemented as a macro-action that (i) executes a short environment interaction sequence designed to reveal $p$, and (ii) parses the resulting observation to produce a truth value $v$ and a set of additional grounded facts $\Delta w$.
We always apply the grounded update:
\begin{equation}
w \leftarrow w \cup \{(p,v)\}\cup \Delta w.
\end{equation}

\paragraph{Budget accounting.}
We follow the benchmark conventions for interaction budgets and replanning limits. Query macro-actions consume environment steps and therefore reduce the remaining episode budget; we do not introduce an additional metric beyond those already reported in prior work.

\section{Reproducibility}
\label{app:engineering}

This appendix collects \emph{implementation-level} design choices that improve robustness and efficiency but are not required by the AEC definition in the main text.

\subsection{Hypothesis Generation and Precondition Annotation}
\label{app:hypothesis_generation}
We generate a small hypothesis set $H$ per task by prompting an LLM to propose candidate plans and annotate each plan with:
(i) $\Pre(h)$ and (ii) $h.\mathrm{expected}(p)$ for $p\in\Pre(h)$.
AEC treats this component as a black box: any planner that produces candidate hypotheses with explicit preconditions can be substituted.

\subsection{World Model Instantiation and Uncertainty Calibration}
\label{app:world_model_impl}
AEC only assumes access to $(\mu_p,\sigma_p)=M_\theta(w,p)$. In our implementation, we instantiate $M_\theta$ with an LLM-based predictor and use lightweight calibration on held-out tasks to map raw confidence signals to $\sigma_p$.
We emphasize that AEC does not rely on a specific uncertainty estimator; alternative schemes (e.g., ensembles, conformal calibration, Bayesian approximations) can be plugged into the same interface.

\subsection{Replanning Trigger}
\label{app:mpc_monitor}
After AEC commits to a plan $h^\star$ (i.e., passes $\mathcal{V}$), we enable an monitor that triggers replanning when grounded counterevidence is observed during execution. We re-enter Algorithm~\ref{alg:aec} when:
(i) an environment action fails unexpectedly, or
(ii) a post-action observation contradicts a grounded precondition required by the remaining plan.
This monitor is conceptually similar to receding-horizon execution (MPC-style) in that it uses observed feedback to decide when to replan, but it does not change the AEC correctness argument: correctness is tied to the verifier-gated commitment step, while monitoring affects empirical robustness.

\subsection{Counterexample-Driven Verifier Repair}
\label{app:verifier_counterexamples}
The main text assumes $\mathcal{V}$ is sound. In practice, a counterexample may indicate that an entailment rule (used by $w\models p$) was too permissive, or that grounding produced a spurious fact.
When a counterexample is detected, we apply the following conservative policy:
\begin{enumerate}
    \item \textbf{Evidence override:} update $w$ with grounded evidence via Eq.~\eqref{eq:query} and discard hypotheses inconsistent with the new grounded facts.
    \item \textbf{Rule tightening (optional):} if the same entailment rule repeatedly produces counterexamples, disable it or add additional guards (type checks / context conditions) to restore soundness.
\end{enumerate}
This policy prioritizes safety: it may increase replanning or query cost, but avoids silently committing to plans based on incorrect inferences.

\end{document}